\documentclass[pdflatex,sn-mathphys-num]{sn-jnl}%
\usepackage{graphicx}%
\usepackage{multirow}%
\usepackage{amsmath,amssymb,amsfonts}%
\usepackage{amsthm}%
\usepackage{mathrsfs}%
\usepackage[title]{appendix}%
\usepackage{xcolor}%
\usepackage{textcomp}%
\usepackage{manyfoot}%
\usepackage{booktabs}%
\usepackage{graphicx}   

\usepackage{algorithmicx}%
\usepackage{algpseudocode}%
\usepackage{listings}%
\usepackage[ruled,vlined]{algorithm2e}
\usepackage{listings}%

\usepackage{subfig}
\usepackage{tikz}
\usetikzlibrary{shapes.geometric, arrows}

\tikzstyle{startstop} = [rectangle, rounded corners, minimum width=3cm, minimum height=1cm,text centered, draw=black, fill=red!30]
\tikzstyle{process} = [rectangle, minimum width=3cm, minimum height=1cm, text centered, draw=black, fill=blue!20]
\tikzstyle{decision} = [diamond, minimum width=3cm, minimum height=1cm, text centered, draw=black, fill=green!30]
\tikzstyle{arrow} = [thick,->,>=stealth]

\begin{document}

\title{Hybrid Topological and Deep Feature Fusion for Accurate MRI-Based Alzheimer’s Disease Severity Classification}

\author*[1]{\fnm{Faisal Ahmed} }\email{ahmedf9@erau.edu}

\affil*[1]{\orgdiv{Department of Data Science and Mathematics}, \orgname{Embry-Riddle Aeronautical University}, \orgaddress{\street{3700 Willow Creek Rd}, \city{Prescott}, \postcode{86301}, \state{Arizona}, \country{USA}}}

\abstract{
Early and accurate diagnosis of Alzheimer’s disease (AD) remains a critical challenge in neuroimaging-based clinical decision support systems. In this work, we propose a novel hybrid deep learning framework that integrates Topological Data Analysis (TDA) with a DenseNet121 backbone for four-class Alzheimer’s disease classification using structural MRI data from the OASIS dataset. TDA is employed to capture complementary topological characteristics of brain structures that are often overlooked by conventional neural networks, while DenseNet121 efficiently learns hierarchical spatial features from MRI slices. The extracted deep and topological features are fused to enhance class separability across the four AD stages. Extensive experiments conducted on the OASIS-1 Kaggle MRI dataset  demonstrate that the proposed TDA+DenseNet121 model significantly outperforms existing state-of-the-art approaches. The model achieves an accuracy of 99.93\% and an AUC of 100\%, surpassing recently published CNN-based, transfer learning, ensemble, and multi-scale architectures. These results confirm the effectiveness of incorporating topological insights into deep learning pipelines and highlight the potential of the proposed framework as a robust and highly accurate tool for automated Alzheimer’s disease diagnosis.

}




\maketitle

\section{Introduction}\label{sec1}

Alzheimer’s disease (AD) is a progressive neurodegenerative disorder and one of the most prevalent causes of dementia worldwide. It is characterized by irreversible cognitive decline accompanied by gradual structural brain degeneration. Accurate identification of AD severity stages is essential for early diagnosis, treatment planning, and monitoring disease progression. Magnetic Resonance Imaging (MRI) has become a cornerstone modality for Alzheimer’s disease assessment due to its non-invasive nature and its ability to capture structural alterations such as hippocampal atrophy, cortical thinning, and ventricular enlargement~\cite{jack2008alzheimer, mosconi2005early}.

Despite its clinical importance, automated MRI-based Alzheimer’s disease classification remains a challenging task. Structural changes across disease stages are often subtle, non-linear, and highly variable across subjects. Moreover, the limited availability of well-annotated neuroimaging datasets and class imbalance issues further complicate reliable model training~\cite{liu2021deep}. Traditional machine learning approaches rely on handcrafted features, which may fail to capture complex spatial and structural relationships inherent in brain MRI data.

Recent advances in deep learning, particularly convolutional neural networks (CNNs), have shown promising results for Alzheimer’s disease classification from MRI~\cite{zhang2021multi}. Architectures such as AlexNet, Siamese networks, multi-scale CNNs, and ensemble-based deep models have been explored on the OASIS and OASIS-derived datasets, achieving high classification accuracy~\cite{maqsood2019transfer, islam2018brain, Siamese2023FourWay, Femmam2024MRI}. However, as summarized in Table~\ref{tab:results}, these models often rely on aggressive data augmentation, large training sets, or complex ensemble strategies, and they primarily learn local intensity-based features. Such approaches can be computationally expensive, sensitive to data imbalance, and difficult to interpret, limiting their clinical applicability~\cite{hernandez2019ensemble}.

Topological Data Analysis (TDA) has recently emerged as a complementary mathematical framework capable of capturing global and shape-aware properties of complex data~\cite{carlsson2009topology}. By leveraging persistent homology, TDA encodes stable topological features that summarize intrinsic structural patterns while remaining robust to noise and small perturbations~\cite{edelsbrunner2008persistent}. In medical imaging, TDA has demonstrated strong potential for modeling anatomical structures and disease-related variations in an interpretable and data-efficient manner~\cite{ahmed2023topo}. However, purely topology-driven models may struggle to capture fine-grained local textures present in high-dimensional MRI data.

Motivated by these observations, we propose a novel hybrid framework that integrates \textbf{Topological Data Analysis with DenseNet121}, combining the strengths of global topological descriptors and deep convolutional feature learning. DenseNet121 is a highly efficient deep architecture that promotes feature reuse through dense connectivity, enabling strong representation learning with fewer parameters~\cite{huang2017densely}. In the proposed model, TDA-based features capture global structural and geometric characteristics of brain MRI, while DenseNet121 learns hierarchical spatial representations, resulting in enhanced feature complementarity.

\begin{figure*}[t!]
    \centering
    \subfloat[\scriptsize 3D PCA projection of Betti-0 features showing clear separation among the four Alzheimer’s disease classes.\label{fig:pca-B0}]{
        \includegraphics[width=0.45\linewidth]{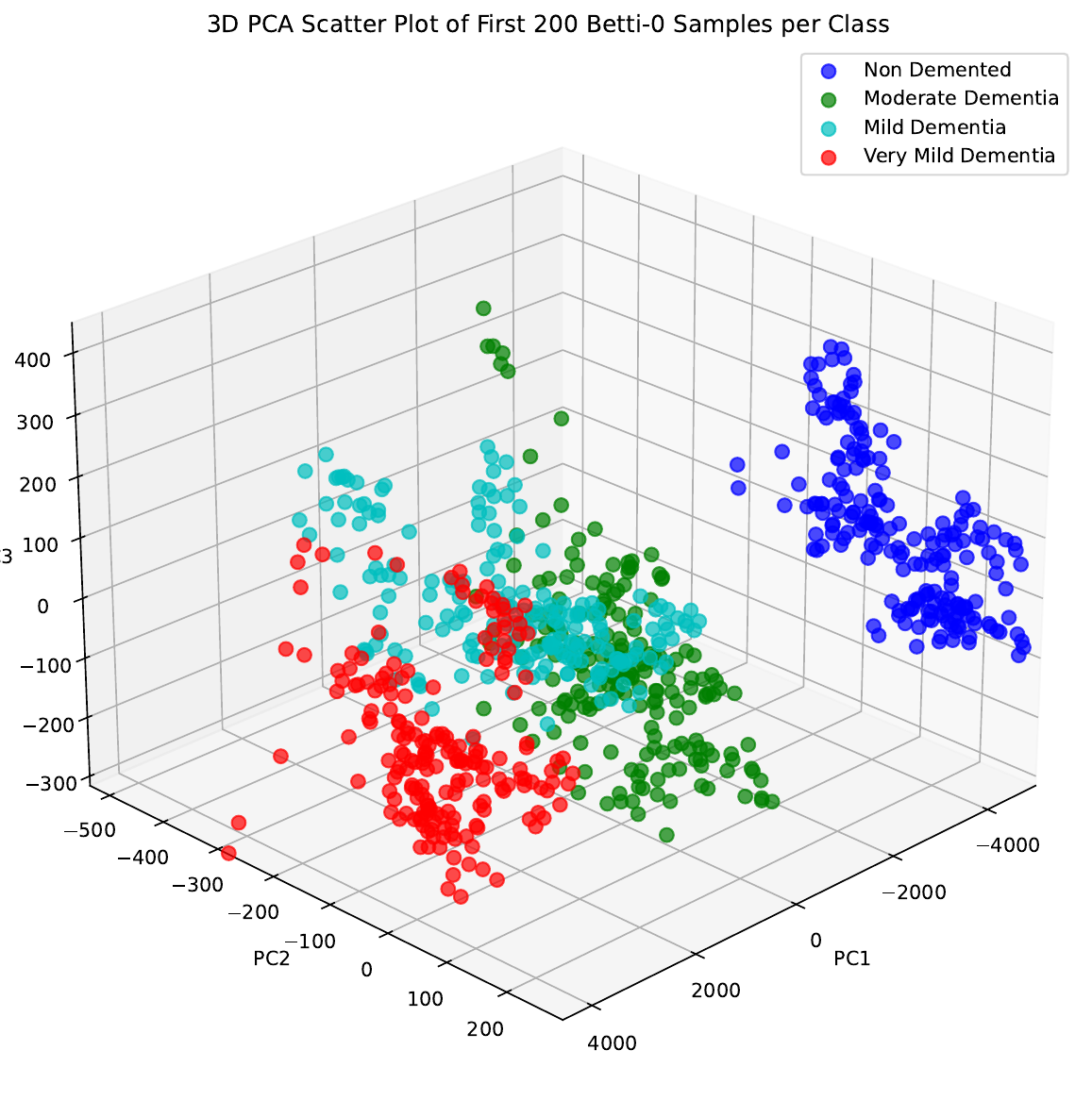}}
    \hfill
    \subfloat[\scriptsize 3D PCA projection of Betti-1 features illustrating distinct clustering of each Alzheimer’s disease class.\label{fig:pca-B1}]{
        \includegraphics[width=0.45\linewidth]{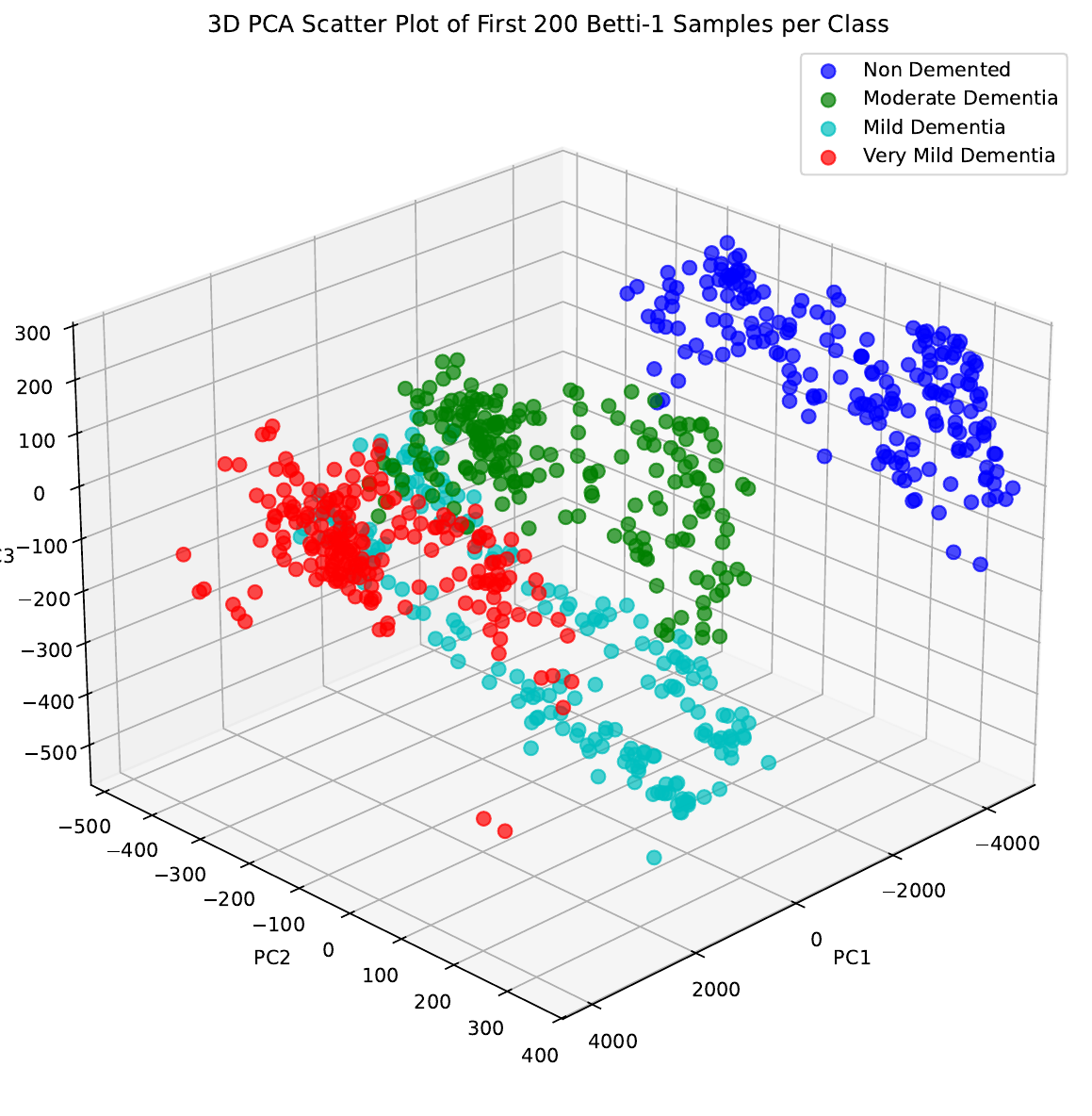}}
    
    \caption{\footnotesize Visual analysis of topological features using the first three principal components. (a) Betti-0 features and (b) Betti-1 features demonstrate clear and distinct clustering corresponding to the four AD severity stages, highlighting the discriminative power of topological descriptors.}
    \label{fig:pca-visualization}
\end{figure*}

\begin{figure*}[t!]
    \centering
    \subfloat[\scriptsize Violin plot of Betti-0 (connected components) features across the four Alzheimer’s disease stages, revealing clearly separated and class-specific distributional patterns.\label{fig:vio-B0}]{
        \includegraphics[width=0.45\linewidth]{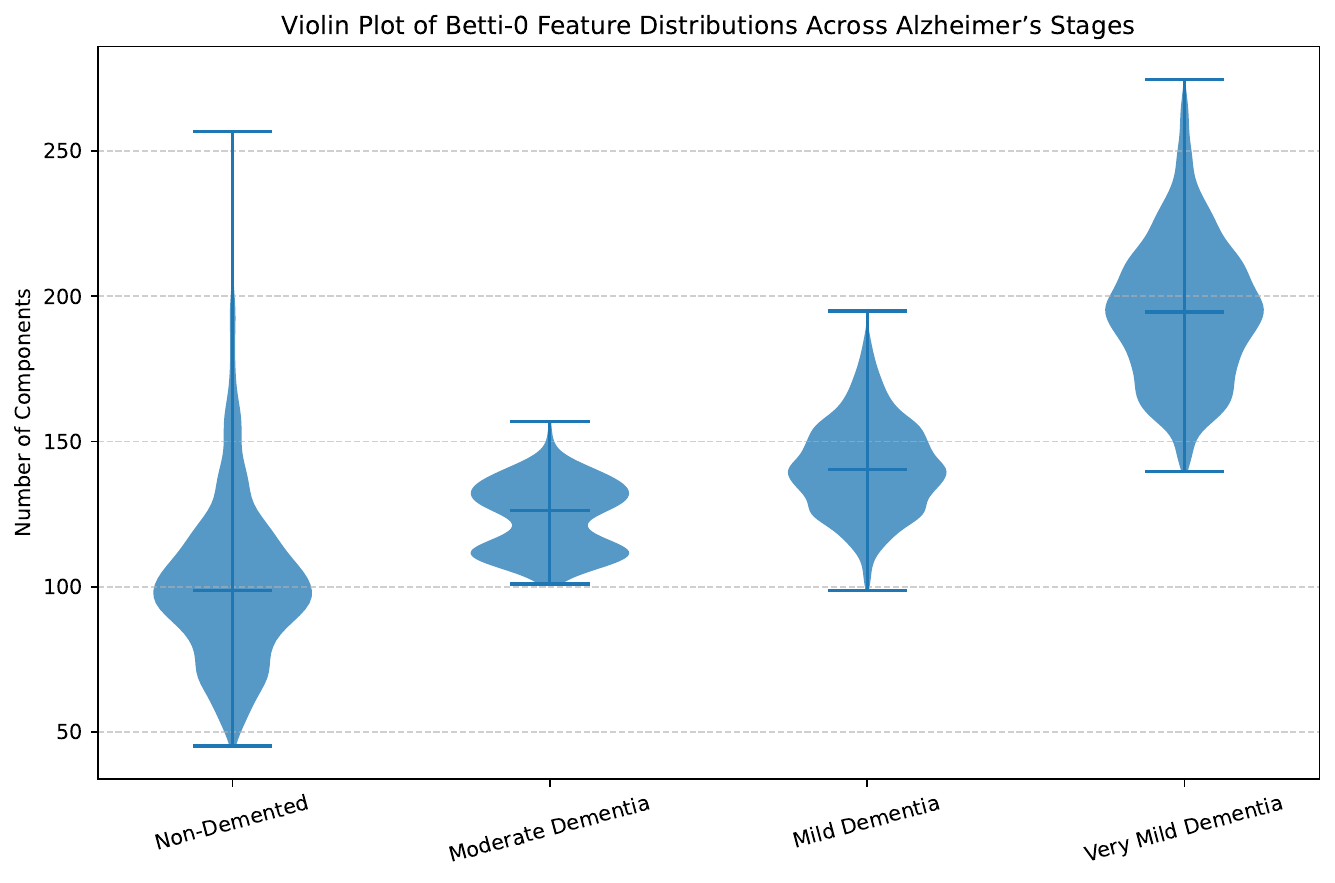}}
    \hfill
    \subfloat[\scriptsize Violin plot of Betti-1 (topological loops) features showing distinct, well-structured distributions for each Alzheimer’s disease class.\label{fig:vio-B1}]{
        \includegraphics[width=0.45\linewidth]{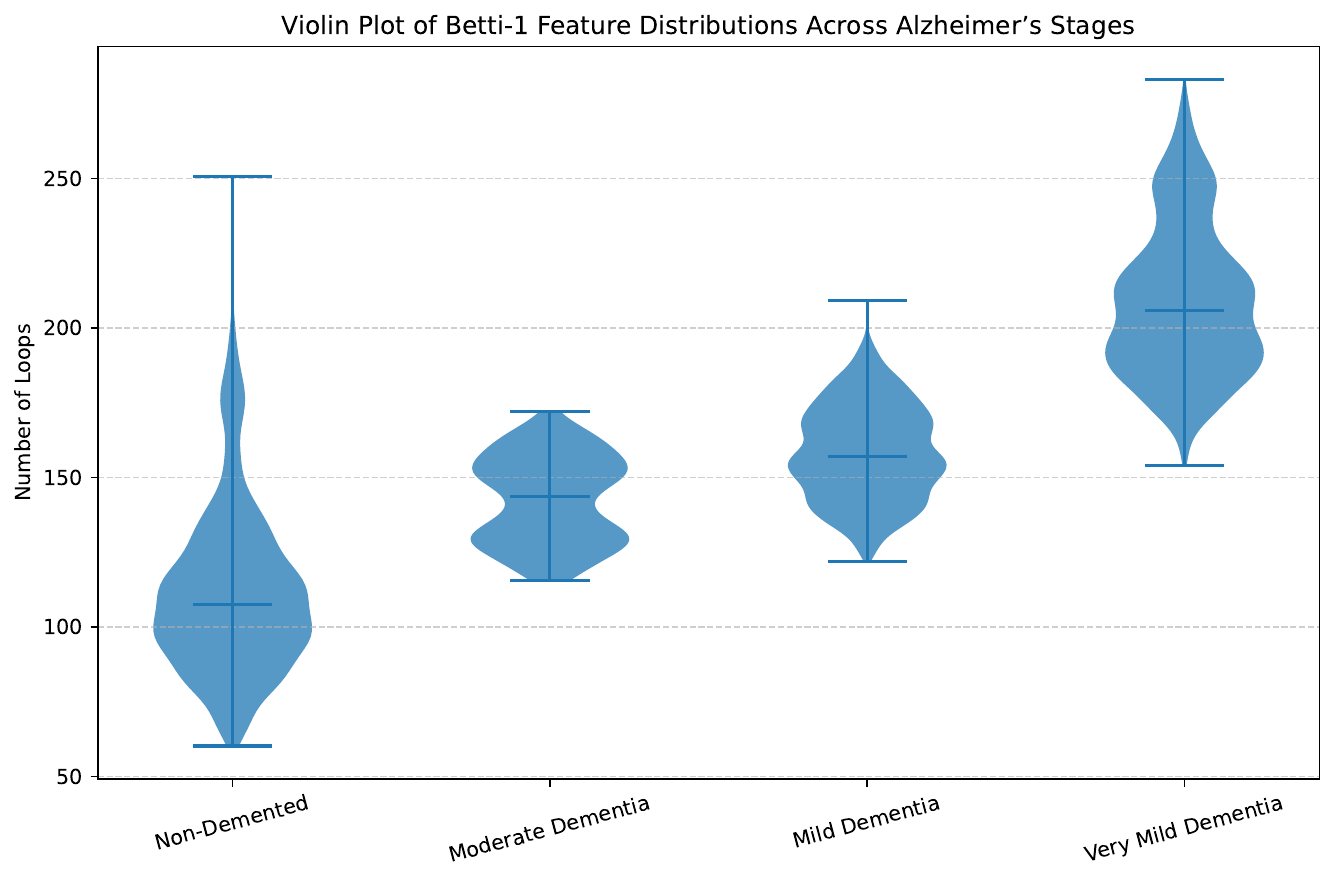}}
    
    \caption{\footnotesize 
    Interpretable visualization of topological feature distributions derived from persistent homology. 
    The violin plots illustrate the class-wise distributions of aggregated Betti-0 and Betti-1 features for all MRI samples. 
    Distinct and non-overlapping patterns are observed across the four Alzheimer’s disease severity stages, indicating progressive and discriminative topological changes in brain structure. 
    These clear separations highlight the strong interpretability and class-discriminative capability of the proposed TDA-based descriptors.}
    \label{fig:violin-visualization}
\end{figure*}

We evaluate the proposed \textbf{TDA+DenseNet121} framework on the \textbf{OASIS-1 (Kaggle) MRI dataset} for four-class Alzheimer’s disease severity classification. As shown in Table~\ref{tab:results}, the proposed model achieves an accuracy of \textbf{99.93\%} and an AUC of \textbf{100\%}, outperforming recently published deep learning and ensemble-based methods on OASIS and OASIS-derived datasets. These results underscore the value of topological representations for Alzheimer’s disease severity assessment and show that the proposed \textbf{TDA+DenseNet121} framework offers a computationally efficient, data-efficient, and interpretable alternative to conventional deep learning–based MRI classification approaches (see Figures~\ref{fig:pca-visualization} and~\ref{fig:violin-visualization}). Moreover, integrating topological information with deep learning substantially enhances model robustness, generalization capability, and overall classification performance.

\noindent\textbf{Our Contributions}
\begin{itemize}
    \item We propose a novel hybrid interpretable  \textbf{ TDA+DenseNet121} framework for four-class Alzheimer’s disease severity classification from structural MRI.
    
    \item We integrate \textbf{topological descriptors derived from persistent homology} with deep convolutional features to capture both global structural patterns and local spatial characteristics of brain MRI.
    
    \item We demonstrate \textbf{state-of-the-art performance} on the OASIS-1 MRI dataset, achieving \textbf{99.93\% accuracy} and \textbf{100\% AUC}, surpassing existing deep learning and ensemble-based approaches.
    
    \item We show that the proposed model provides improved \textbf{robustness and generalization} while maintaining computational efficiency, making it suitable for clinical and limited-data scenarios.
    
    \item We highlight the potential of \textbf{topology-aware deep learning} as an effective and interpretable paradigm for neuroimaging-based Alzheimer’s disease diagnosis.
\end{itemize}

\begin{figure}[t!]
	\centering
	\subfloat[\scriptsize Non-demented MRI sample.\label{fig:ND}]{%
		\includegraphics[width=0.2\linewidth]{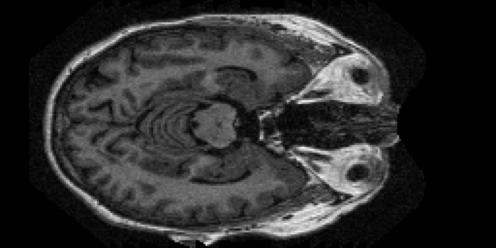}}
	\hfill
	\subfloat[\scriptsize Moderate dementia MRI sample.\label{fig:MD}]{%
		\includegraphics[width=0.2\linewidth]{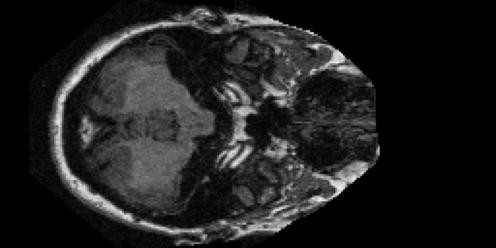}}
	\hfill
	\subfloat[\scriptsize Mild dementia MRI sample.\label{fig:MiD}]{%
		\includegraphics[width=0.2\linewidth]{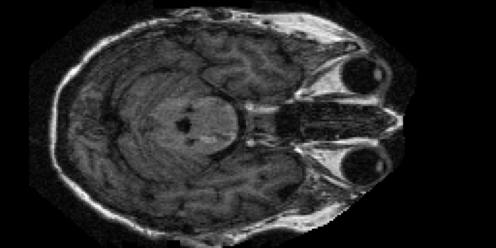}}
    \hfill
	\subfloat[\scriptsize Very mild dementia MRI sample.\label{fig:VMD}]{%
		\includegraphics[width=0.2\linewidth]{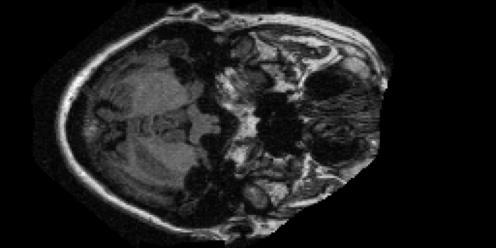}}

	\caption{\footnotesize Representative brain MRI examples from the OASIS-1 dataset corresponding to the four Alzheimer’s disease categories considered in this study.}
	\label{fig:image-samples}
\end{figure}

\section{Related Works}\label{sec2}

Automated analysis of brain magnetic resonance imaging (MRI) has been extensively studied for computer-aided diagnosis of Alzheimer’s disease (AD). Early research primarily relied on handcrafted features derived from voxel-based morphometry (VBM), cortical thickness, and regional volumetric measurements, combined with conventional machine learning classifiers. Support Vector Machines (SVMs) demonstrated promising results in distinguishing Alzheimer’s patients from cognitively normal subjects by exploiting structural brain alterations~\cite{Kloppel2008SVM}. Subsequent studies integrated VBM features with convolutional neural networks (CNNs) to enhance discriminative power, although these methods remained sensitive to preprocessing pipelines and dataset-specific variations~\cite{Zhang2022VBMCNN}.

The rapid development of deep learning significantly advanced MRI-based AD classification. Both 2D and 3D CNN architectures have been widely adopted due to their ability to automatically learn hierarchical representations from neuroimaging data~\cite{ebrahimi2021convolutional}. Dense and multi-scale CNN architectures further improved classification accuracy by capturing complementary local and contextual information~\cite{wang2021densecnn}. Ensemble learning strategies combining multiple CNN models have also been explored to enhance robustness and generalization~\cite{fathi2024deep}. Despite their strong performance, these approaches typically depend on large annotated datasets, extensive data augmentation, and high computational cost, which limit their applicability in small or moderately sized clinical datasets.

To mitigate the limitations of purely deep learning-based models, hybrid approaches incorporating structural priors and feature selection techniques have been proposed. Topology-preserving segmentation frameworks, such as TOADS, enforce anatomical consistency during brain tissue segmentation~\cite{bazin2007topology}. Feature selection methods, including minimum Redundancy Maximum Relevance (mRMR), have been applied to reduce feature dimensionality and improve classification stability in MRI-based AD studies~\cite{alshamlan2023identifying, alshamlan2024improving}. While these techniques enhance interpretability and robustness, they largely rely on intensity-based or region-specific features and may not fully capture global structural changes associated with disease progression.

Recently, Topological Data Analysis (TDA) has emerged as a powerful framework for extracting global, shape-aware representations from complex medical imaging data. By leveraging persistent homology, TDA encodes intrinsic topological characteristics that remain stable under noise and small perturbations~\cite{carlsson2009topology, edelsbrunner2008persistent}. A growing body of research has demonstrated the effectiveness of TDA in medical image analysis, including applications to brain MRI, histopathology, and volumetric imaging~\cite{ahmed2026four, ahmed2025topo, ahmed2023topo, ahmed2023topological, ahmed2023tofi, ahmed20253d, yadav2023histopathological, ahmed2025topological}. These studies highlight the robustness, interpretability, and data efficiency of topological descriptors, particularly in limited-data scenarios.

In parallel, Vision Transformers (ViTs) have gained attention as an alternative to CNNs by leveraging self-attention mechanisms to model long-range dependencies~\cite{dosovitskiy2021image, liu2021swin}. ViT-based models have shown promising performance in medical imaging tasks, including brain MRI analysis~\cite{sankari2025hierarchical, dhinagar2023efficiently}. However, their application to grayscale MRI data remains challenging due to high computational requirements, reliance on large-scale pretraining, and adaptations such as channel replication, which may not fully exploit MRI-specific structural information. Recent studies have explored transfer learning and transformer-based architectures for medical image analysis, yet performance often degrades in small or imbalanced datasets~\cite{ahmed2025colormap, ahmed2025hog, ahmed2025ocuvit, ahmed2025robust, ahmed2025histovit, ahmed2025addressing, ahmed2025repvit, ahmed2025pseudocolorvit, rawat2025efficient, ahmed2025transfer}.

In contrast, topology-aware approaches offer a computationally efficient and interpretable alternative for MRI-based Alzheimer’s disease classification. By capturing global structural patterns that complement local intensity-based representations, TDA-based methods address key limitations of both deep learning and transformer-based models. Motivated by these advantages, the present work integrates topological descriptors with deep convolutional features to achieve accurate and robust four-stage Alzheimer’s disease severity classification without excessive reliance on data augmentation or large-scale training.

\section{Methodology}\label{sec:method}

\begin{figure*}[t!]
    \centering
    \includegraphics[width=\linewidth]{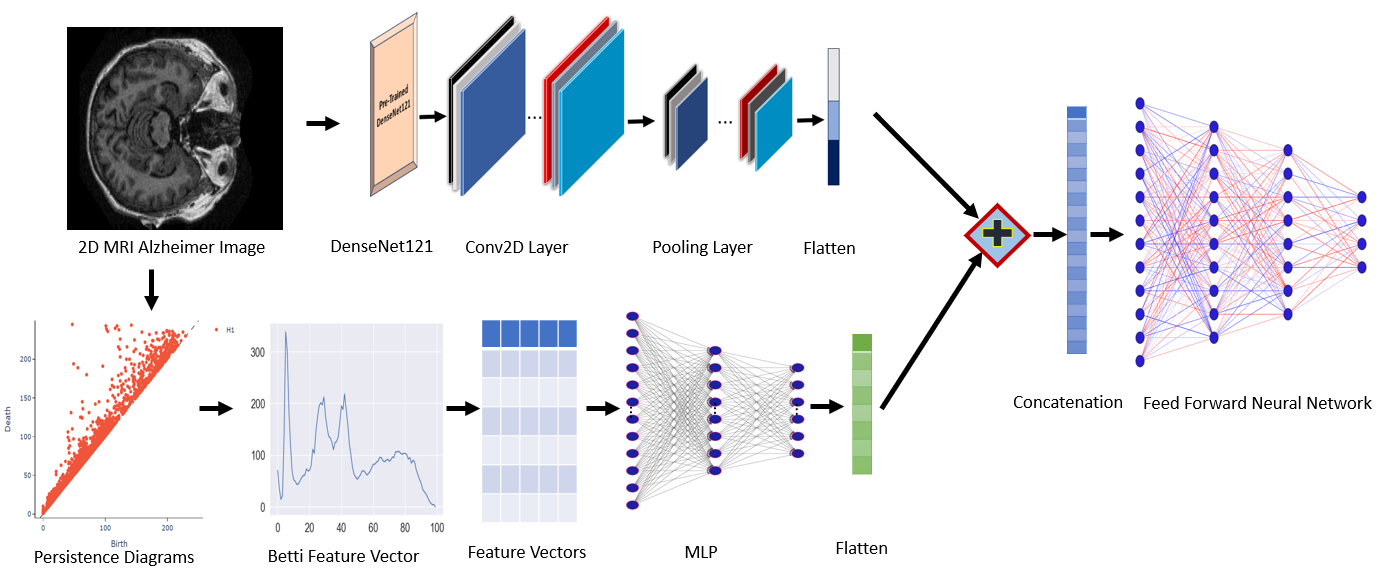}
    \caption{\footnotesize
    \textbf{Overview of the proposed TDA+DenseNet121 framework for Alzheimer’s disease classification.}
    The architecture consists of two complementary feature extraction branches. In the first branch, a 2D brain MRI slice is processed through a DenseNet121-based convolutional network to learn hierarchical deep features, followed by convolution, pooling, and flattening operations. In parallel, the second branch applies Topological Data Analysis (TDA) to the same MRI input, where cubical sublevel filtrations are used to compute persistence diagrams that capture the birth and death of topological features. These diagrams are transformed into fixed-length Betti feature vectors and further refined using a multilayer perceptron (MLP). The deep convolutional features and topological features are then concatenated to form a unified representation, which is finally fed into a feed-forward neural network for four-class Alzheimer’s disease severity classification.
    }
    \label{fig:flowchart}
\end{figure*}

\subsection{Topological Feature Extraction via Persistent Homology}
\label{subsec:tda}

\begin{figure}[t]
    \centering
    \includegraphics[width=\linewidth]{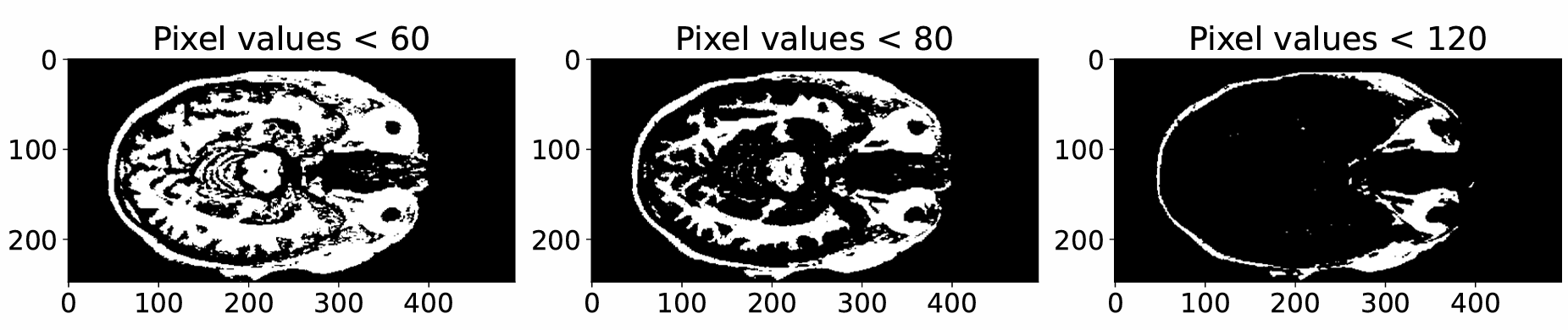}
    \caption{\small 
    \textbf{Sublevel filtration of an MRI image.} 
    Illustration of the cubical (sublevel) filtration process, where pixels are progressively activated based on their grayscale intensity values. The binary images 
    $\mathcal{X}_{60}$, $\mathcal{X}_{80}$, and $\mathcal{X}_{120}$ correspond to threshold values of $60$, $80$, and $120$, respectively, revealing the gradual emergence and evolution of connected components and topological structures across increasing intensity levels.
    }
    \label{fig:filtration}
\end{figure}

Topological data analysis (TDA) is employed to extract robust structural descriptors from 2D MRI slices (see Figure \ref{fig:image-samples}) using persistent homology (PH). Given a grayscale MRI image 
$\mathcal{X} \in \mathbb{R}^{H \times W}$,
a cubical sublevel filtration is constructed by progressively thresholding pixel intensities (see Figure \ref{fig:filtration}). For a sequence of increasing thresholds 
$\{t_1, t_2, \ldots, t_N\}$,
a filtration is constructed as a nested sequence of cubical complexes, which can be interpreted as successive binary images:
\[
\mathcal{X}_{t_1} \subseteq \mathcal{X}_{t_2} \subseteq \cdots \subseteq \mathcal{X}_{t_N},
\quad
\mathcal{X}_{t_k} = \{ (i,j) \mid \mathcal{X}_{i,j} \le t_k \}.
\]

For each filtration level, homology groups $H_k(\mathcal{X}_{t_k})$ are computed, where $k=0$ captures connected components and $k=1$ captures loops or cavities. The birth and death times of these topological features are summarized using persistence diagrams:
\[
\mathrm{PD}_k = \{ (b_\ell, d_\ell) \mid \ell \in H_k \},
\]
where $b_\ell$ and $d_\ell$ denote the threshold values at which a feature appears and disappears, respectively.

To transform persistence diagrams into fixed-length representations suitable for learning, Betti curves are computed. The Betti number $\beta_k(t)$ counts the number of $k$-dimensional topological features active at threshold $t$:
\[
\beta_k(t) = \left| \{ (b_\ell, d_\ell) \in \mathrm{PD}_k \mid b_\ell \le t < d_\ell \} \right|.
\]
For computational efficiency, Betti curves are discretized into 100 bins for each homology dimension. The final topological feature vector for each MRI slice is obtained by concatenating Betti-0 and Betti-1 vectors:
\[
\mathbf{T} = [\boldsymbol{\beta}_0, \boldsymbol{\beta}_1] \in \mathbb{R}^{200}.
\]
These vectors encode global morphological characteristics of brain structures and are invariant to geometric transformations such as translation and rotation.

\subsection{DenseNet121-Based Image Feature Extraction}
\label{subsec:densenet}

To learn discriminative spatial features directly from MRI images, a deep convolutional neural network based on DenseNet121 is employed. Each MRI slice is first normalized and reshaped into a tensor
$\mathbf{I} \in \mathbb{R}^{248 \times 248 \times 1}$.
Since DenseNet121 is designed for three-channel inputs, an initial convolutional layer is applied to project the grayscale image into a three-channel representation:
\[
\mathbf{I}' = \sigma(\mathbf{W}_{\text{conv}} * \mathbf{I} + \mathbf{b}_{\text{conv}}),
\quad
\mathbf{I}' \in \mathbb{R}^{248 \times 248 \times 3},
\]
where $*$ denotes convolution and $\sigma(\cdot)$ is the ReLU activation.

The transformed image $\mathbf{I}'$ is passed through the DenseNet121 backbone with the classification head removed. DenseNet121 consists of densely connected convolutional blocks, where each layer receives feature maps from all preceding layers, facilitating feature reuse and improved gradient flow. Let $\Phi(\cdot)$ denote the mapping learned by the convolutional backbone:
\[
\mathbf{F}_{\text{Dense}} = \Phi(\mathbf{I}'),
\quad
\mathbf{F}_{\text{Dense}} \in \mathbb{R}^{h \times w \times c}.
\]

Global average pooling is applied to aggregate spatial information:
\[
\mathbf{f}_{\text{GAP}} = \frac{1}{hw} \sum_{i=1}^{h} \sum_{j=1}^{w} \mathbf{F}_{\text{Dense}}(i,j),
\quad
\mathbf{f}_{\text{GAP}} \in \mathbb{R}^{c}.
\]
This vector is further refined through a fully connected layer with ReLU activation to produce the image feature embedding:
\[
\mathbf{F}_{\text{CNN}} = \text{ReLU}(\mathbf{W}_{\text{cnn}} \mathbf{f}_{\text{GAP}} + \mathbf{b}_{\text{cnn}}),
\quad
\mathbf{F}_{\text{CNN}} \in \mathbb{R}^{64}.
\]
These features capture local texture patterns and spatial variations relevant to Alzheimer’s disease progression.

\subsection{Feature Fusion and Classification}
\label{subsec:fusion}

The proposed framework integrates topological and deep image features through a late-fusion strategy. The topological embedding
$\mathbf{F}_{\text{TDA}} \in \mathbb{R}^{128}$
and the CNN embedding
$\mathbf{F}_{\text{CNN}} \in \mathbb{R}^{64}$
are concatenated to form a joint feature representation:
\[
\mathbf{F}_{\text{fusion}} = [\mathbf{F}_{\text{TDA}}, \mathbf{F}_{\text{CNN}}] \in \mathbb{R}^{192}.
\]

This fused vector is passed through a sequence of fully connected layers to model higher-order interactions between modalities:
\[
\mathbf{z}_1 = \text{ReLU}(\mathbf{W}_1 \mathbf{F}_{\text{fusion}} + \mathbf{b}_1),
\quad
\mathbf{z}_2 = \text{ReLU}(\mathbf{W}_2 \mathbf{z}_1 + \mathbf{b}_2).
\]
A dropout operation with rate $p=0.2$ is applied to reduce overfitting. The final classification layer uses a softmax activation to predict the probability distribution over four Alzheimer’s disease stages:
\[
\hat{\mathbf{y}} = \text{Softmax}(\mathbf{W}_{\text{out}} \mathbf{z}_2 + \mathbf{b}_{\text{out}}),
\quad
\hat{\mathbf{y}} \in \mathbb{R}^{4}.
\]

The model is trained by minimizing the categorical cross-entropy loss:
\[
\mathcal{L} = - \sum_{c=1}^{4} y_c \log(\hat{y}_c),
\]
where $y_c$ denotes the ground-truth label. The fusion strategy enables the model to jointly exploit global topological structure and local spatial patterns, resulting in improved classification performance. The complete workflow of the proposed model is illustrated in Figure~\ref{fig:flowchart}. The algorithmic framework of the proposed method is presented in Algorithm~\ref{alg:tda_densenet}.

\subsection{Hyperparameter Tuning}
\label{subsec:hyperparameter}

Hyperparameters for the proposed TDA+DenseNet121 fusion model were selected empirically and fixed across all experiments to ensure reproducibility. The model was implemented in TensorFlow/Keras and trained end-to-end.The dataset was divided into training and testing subsets using a $90{:}10$ split with a fixed random seed. Class labels were encoded using one-hot encoding for four-class classification. MRI images were resized to $248 \times 248$ pixels and normalized to the range $[0,1]$ prior to training. The DenseNet121 branch was initialized with ImageNet pretrained weights and configured without the classification head. Global average pooling was applied to the convolutional feature maps, followed by a fully connected layer with ReLU activation. In parallel, the topological feature branch employed a multilayer perceptron to process the extracted TDA vectors.

Feature-level fusion was performed by concatenating the outputs of the DenseNet121 and TDA branches, followed by fully connected layers and dropout to mitigate overfitting. The final prediction layer used a softmax activation for four-class classification. Model training was conducted using the Adam optimizer with categorical cross-entropy loss. Training proceeded for 50 epochs with a batch size of 32. Model checkpointing was applied based on validation accuracy, and the best-performing model was retained for evaluation. No data augmentation or learning rate scheduling was used. The complete set of hyperparameter values is summarized in Table~\ref{tab:hyperparameters}.

\begin{table}[h!]
\centering
\caption{Hyperparameter Settings for the Proposed TDA+DenseNet121 Fusion Model}
\label{tab:hyperparameters}
\setlength{\tabcolsep}{6pt}
\footnotesize
\begin{tabular}{ll}
\toprule
\textbf{Component} & \textbf{Hyperparameter Setting} \\
\midrule
\multirow{4}{*}{Input Data} 
& Image size: $248 \times 248 \times 1$ (grayscale) \\
& Pixel normalization: $[0,1]$ scaling \\
& TDA feature dimension: 198 \\
& Train--test split: $90{:}10$ (random seed = 3) \\
\midrule
\multirow{4}{*}{DenseNet121 Branch} 
& Backbone: DenseNet121 \\
& Pretrained weights: ImageNet \\
& Include top layers: False \\
& Dense layer: 64 neurons (ReLU) \\
\midrule
\multirow{4}{*}{TDA-MLP Branch} 
& Fully connected layers: 800--256--128 \\
& Activation function: ReLU \\
& Input dimension: 200 \\
& Output embedding size: 128 \\
\midrule
\multirow{4}{*}{Fusion Network} 
& Fusion method: Feature concatenation \\
& Dense layers: 256, 128 (ReLU) \\
& Dropout rate: 0.2 \\
& Output layer: 4 neurons (Softmax) \\
\midrule
\multirow{5}{*}{Training Setup} 
& Optimizer: Adam (default parameters) \\
& Loss function: Categorical cross-entropy \\
& Batch size: 32 \\
& Epochs: 50 \\
& Model checkpoint: Best validation accuracy \\
\bottomrule
\end{tabular}
\end{table}

\begin{algorithm}[htbp]
\small
\SetAlgoNlRelativeSize{0}
\DontPrintSemicolon
\caption{TDA+DenseNet121: Fusion-Based Four-Class Alzheimer’s Disease Classification}
\label{alg:tda_densenet}

\KwIn{
MRI dataset $\mathcal{D}=\{(\mathbf{I}_i, y_i)\}_{i=1}^{N}$,
homology dimensions $k \in \{0,1\}$,
number of Betti bins $B=100$,
train--test split ratio,
DenseNet121 backbone
}
\KwOut{
Trained fusion model and evaluation metrics
}

\textbf{Step 1: MRI Preprocessing} \\
\For{$i \gets 1$ \KwTo $N$}{
Resize MRI slice $\mathbf{I}_i$ to $248 \times 248$ \;
Normalize pixel intensities to $[0,1]$ \;
Expand channel dimension: $\mathbf{I}_i \in \mathbb{R}^{248 \times 248 \times 1}$ \;
Assign class label $y_i \in \{0,1,2,3\}$ \;
}

\textbf{Step 2: Topological Feature Extraction via Persistent Homology} \\
\For{$i \gets 1$ \KwTo $N$}{
Construct cubical sublevel filtration from $\mathbf{I}_i$ \;
\For{homology dimension $k \in \{0,1\}$}{
Compute persistence diagram
\[
\mathrm{PD}_k(\mathbf{I}_i) = \{(b_\sigma, d_\sigma)\}
\]
Compute Betti curve
\[
\vec{\beta}_k(\mathbf{I}_i) = [\beta_k(t_1), \dots, \beta_k(t_B)]
\]
}
Concatenate Betti features:
\[
\mathbf{T}_i = [\vec{\beta}_0(\mathbf{I}_i), \vec{\beta}_1(\mathbf{I}_i)] \in \mathbb{R}^{200}
\]
}

\textbf{Step 3: Dataset Construction and Split} \\
Construct MRI tensor dataset $\mathbf{X}_{\text{MRI}}$ and TDA feature matrix $\mathbf{X}_{\text{TDA}}$ \;
Split data into training and testing sets using a $90{:}10$ ratio \;
Apply one-hot encoding to class labels \;

\textbf{Step 4: Feature Learning via Dual-Branch Network} \\
\textbf{(a) TDA Branch (MLP):} \\
Process $\mathbf{T}_i$ through fully connected layers:
\[
198 \rightarrow 800 \rightarrow 256 \rightarrow 128
\]
using ReLU activations to obtain topological embedding $\mathbf{F}_{\text{TDA}}$ \;

\textbf{(b) CNN Branch (DenseNet121):} \\
Convert grayscale MRI to 3-channel input \;
Extract deep features using DenseNet121 (ImageNet weights, no classification head) \;
Apply global average pooling and a 64-unit dense layer to obtain image embedding $\mathbf{F}_{\text{CNN}}$ \;

\textbf{Step 5: Feature Fusion and Classification} \\
Concatenate feature embeddings:
\[
\mathbf{F}_{\text{fusion}} = [\mathbf{F}_{\text{TDA}}, \mathbf{F}_{\text{CNN}}]
\]
Pass through dense layers $256 \rightarrow 128$ with ReLU activation \;
Apply dropout with rate $p=0.2$ \;
Predict class probabilities using softmax:
\[
\hat{\mathbf{y}} = \text{Softmax}(\mathbf{W}\mathbf{F}_{\text{fusion}} + \mathbf{b})
\]

\textbf{Step 6: Model Training} \\
Optimize categorical cross-entropy loss using Adam optimizer \;
Train for 50 epochs with batch size 32 \;
Save model with highest validation accuracy using checkpointing \;

\textbf{Step 7: Evaluation} \\
Generate predictions on test set \;
Compute accuracy, precision, recall, F1-score \;
Compute multi-class ROC-AUC (OvR) \;
Generate confusion matrix \;

\Return Trained fusion model and performance metrics
\end{algorithm}

\section{Experiment}

\subsection{Datasets}


\noindent \textbf{OASIS-1 Dataset}~\cite{Marcus2007OpenAccess}

The OASIS-1 dataset is a widely used cross-sectional structural MRI repository comprising T1-weighted brain scans from 416 adult subjects aged between 18 and 96 years. For each subject, three or four T1-weighted MRI scans were acquired during a single imaging session, enabling the generation of high signal-to-noise ratio averaged images and facilitating robust morphometric analysis. Among the total cohort, 100 subjects aged over 60 years were clinically diagnosed with very mild to moderate Alzheimer’s disease (AD), while the remaining 316 subjects were classified as cognitively normal controls. Owing to the multiple scans per subject, the dataset contains approximately 300–400 AD scans and 950–1300 nondemented control scans.

The dataset also includes a reliability subset consisting of 20 nondemented subjects who were rescanned within approximately 90 days, allowing evaluation of test--retest reproducibility. In addition to raw T1-weighted images, OASIS-1 provides motion-corrected and averaged volumes, atlas-registered images, gain-field corrected scans, and skull-stripped brain volumes. Segmentation outputs delineating gray matter, white matter, and cerebrospinal fluid are also available. Furthermore, comprehensive metadata accompany each subject, including demographic information (age, sex, handedness), clinical dementia ratings, and volumetric measures such as estimated total intracranial volume, normalized whole-brain volume, and atlas scaling factors. Due to its standardized acquisition protocol, high data quality, and inclusion of both healthy and Alzheimer’s disease subjects, OASIS-1 has become a benchmark dataset for aging, neurodegenerative disease research, and algorithm validation.

In this study, the OASIS-1 MRI data were organized into four diagnostic categories: nondemented, very mild dementia, mild dementia, and moderate dementia. To ensure robust model training and evaluation, class-specific subsets were constructed from the available scans. The final dataset used for model development comprised 5000 nondemented samples, 5000 very mild dementia samples, 5002 mild dementia samples, and 488 moderate dementia samples. This distribution reflects the intrinsic class imbalance present in the original cohort, particularly the limited availability of moderate dementia cases. The larger sample sizes for nondemented and early-stage dementia classes enabled the model to effectively learn subtle structural changes associated with early Alzheimer’s disease progression, while preserving clinically meaningful diagnostic diversity across all four stages.

\subsection{Experimental Setup}
\noindent \textbf{Training–Test Split:} Following common practice in the literature for OASIS-based Alzheimer’s disease classification, the dataset is partitioned into training and testing subsets using an 90:10 ratio.
\smallskip

\noindent \textbf{No Data Augmentation:} 
In contrast to conventional CNN-based and deep learning approaches that depend heavily on extensive data augmentation strategies to mitigate limited or imbalanced training data~\cite{goutam2022comprehensive}, the proposed \textbf{TDA+DenseNet121} framework does not require any form of data augmentation. The extracted topological descriptors are inherently invariant to geometric transformations such as rotation and flipping, as well as to minor intensity variations, making the model robust to noise and small perturbations in MRI scans. This eliminates the need for augmentation, significantly reduces computational overhead, and improves training efficiency while maintaining strong generalization performance.


\noindent \textbf{Runtime Efficiency and Platform:} 
All experiments were conducted on a personal laptop equipped with an Apple M1 system-on-chip, featuring an 8-core CPU (4 performance cores and 4 efficiency cores) and 16~GB of unified memory. The implementation was carried out in Python using the TensorFlow/Keras framework, and all models were trained and evaluated on macOS. The source code will be made publicly available to support reproducibility.

\section{Results}\label{sec:results}

Table~\ref{tab:results} presents a comparative analysis of recently published methods for four-class Alzheimer’s disease classification using the OASIS and OASIS-derived MRI datasets. The reported approaches span a wide range of deep learning paradigms, including conventional CNNs with extensive data augmentation, transfer learning, ensemble-based architectures, Siamese networks, and multi-scale convolutional models. Performance is evaluated primarily in terms of classification accuracy and, where available, the area under the ROC curve (AUC).

Early transfer learning and ensemble-based approaches achieved moderate performance on the OASIS dataset. The AlexNet-based transfer learning model reported an accuracy of 92.85\% using an 80:20 train--test split~\cite{maqsood2019transfer}, while an ensemble of deep neural networks achieved 93.18\% accuracy under a 90:10 split~\cite{islam2018brain}. Although these methods demonstrated the feasibility of multi-class AD classification, their performance remained limited, likely due to insufficient modeling of subtle inter-class structural differences.

More recent architectures introduced specialized network designs to improve discriminative capability. The four-way Siamese CNN achieved an accuracy of 93.85\% with an AUC of 95.10\% on the OASIS-3 dataset~\cite{Siamese2023FourWay}, highlighting the benefit of metric learning for multi-class discrimination. Similarly, the deep multi-scale CNN model reported improved performance, achieving 98.00\% accuracy and 99.33\% AUC using a 90:10 split~\cite{Femmam2024MRI}. These results demonstrate that multi-scale feature extraction enhances the capture of disease-related patterns; however, such models typically rely on complex architectures and increased computational cost.

The highest performance among prior works was reported by CNN-based models employing extensive data augmentation, achieving up to 99.68\% accuracy on the OASIS Kaggle MRI dataset using a 70:30 split~\cite{Dardouri2025EfficientCNN}. While highly effective, these approaches depend heavily on aggressive augmentation strategies to mitigate data scarcity and improve generalization, which may introduce redundancy and limit reproducibility across datasets.

In contrast, the proposed \textbf{TDA+DenseNet121} framework achieves the best overall performance, attaining an accuracy of \textbf{99.93\%} and an AUC of \textbf{100\%} on the OASIS-1 Kaggle MRI dataset using a 90:10 train--test split. This improvement is achieved without reliance on excessive data augmentation or complex ensemble mechanisms. The superior performance can be attributed to the complementary integration of topological descriptors, which capture global structural and geometric properties of brain MRI, with deep convolutional features learned by DenseNet121. This hybrid representation enhances class separability across all four Alzheimer’s disease stages.

Overall, the results demonstrate that incorporating Topological Data Analysis into deep learning pipelines provides a robust and highly discriminative feature representation for Alzheimer’s disease severity classification. Compared with existing state-of-the-art CNN, Siamese, and ensemble-based approaches, the proposed method offers improved accuracy, excellent AUC performance, and greater robustness, highlighting its potential for reliable and clinically relevant MRI-based Alzheimer’s disease diagnosis.

\begin{figure*}[t!]
    \centering
    \subfloat[\scriptsize One-vs-rest ROC curves with corresponding AUC scores for each class.\label{fig:auc}]{
        \includegraphics[width=0.45\linewidth]{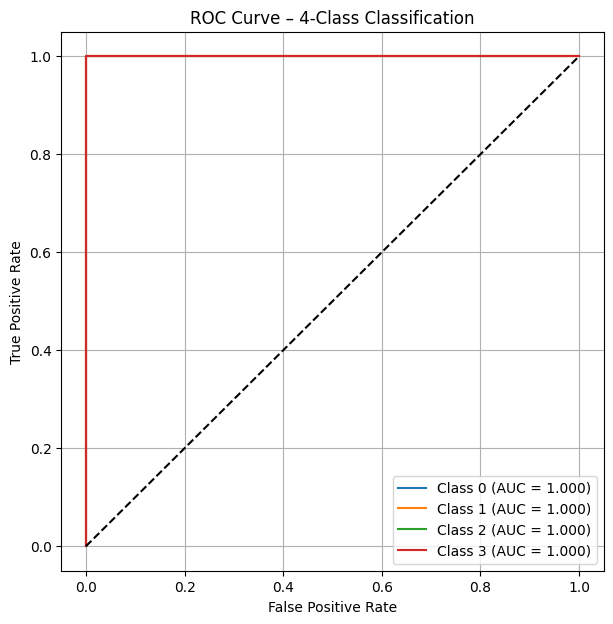}}
    \hfill
    \subfloat[\scriptsize Confusion matrix showing class-wise prediction performance for four-class Alzheimer’s disease classification.\label{fig:conf}]{
        \includegraphics[width=0.45\linewidth]{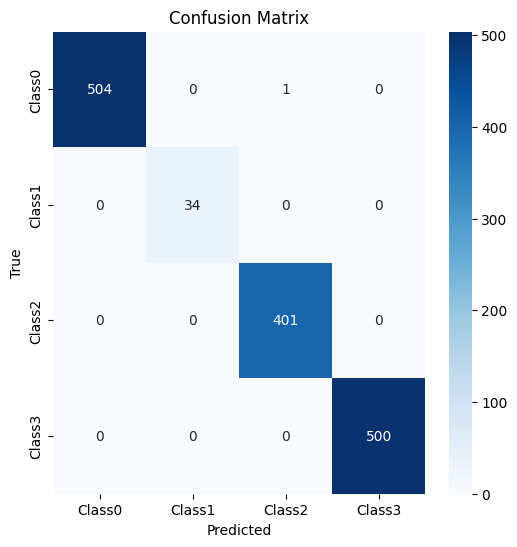}}
    
    \caption{\footnotesize Evaluation of \textbf{TDA+DenseNet121} on the OASIS-1 dataset. (a) One-vs-rest ROC curves illustrating discrimination performance across the four Alzheimer’s disease classes. (b) Confusion matrix showing correct and misclassified instances for each class.}
    \label{fig:auc_conf}
\end{figure*}

\begin{table}[h!]
\centering
\caption{\footnotesize 
Published accuracy results for four-class Alzheimer’s disease classification 
on OASIS or OASIS-derived MRI datasets. 
\label{tab:results}}
\setlength\tabcolsep{4pt}
\footnotesize

\begin{tabular}{lccccc}
\multicolumn{6}{c}{\bf OASIS / OASIS-derived MRI Dataset: 4-Class Classification Results} \\
\toprule
Method & \# Classes & Dataset & Train:Test & Accuracy & AUC \\
\midrule

CNN (DA + augmentation)~\cite{Dardouri2025EfficientCNN} 
& 4 & OASIS (Kaggle MRI) & 70:30 & 99.68 & -- \\

Transfer Learning using AlexNet ~\cite{maqsood2019transfer} 
& 4 & OASIS MRI & 80:20 & 92.85 & -- \\

Ensemble of deep neural network~\cite{islam2018brain} 
& 4 & OASIS MRI & 90:10 & 93.18 & -- \\

Four-way Siamese CNN~\cite{Siamese2023FourWay} 
& 4 & OASIS-3 MRI & 80:20 & 93.85 & 95.10 \\
Deep Multi-scale CNN~\cite{Femmam2024MRI} 
& 4 & OASIS (MRI) & 90:10 & 98.00 & 99.33 \\
CNN (DA + augmentation)~\cite{Dardouri2025EfficientCNN} 
& 4 & OASIS (Kaggle MRI) & 70:30 & 99.68\% & -- \\

\midrule
\bf TDA+DenseNet121(Ours) 
& 4 & OASIS-1(Kaggle MRI) & 90:10 & \textbf{99.93} & \textbf{100} \\
\bottomrule
\end{tabular}
\end{table}

\section{Discussion}\label{sec:discussion}

The results presented in this study demonstrate that integrating Topological Data Analysis (TDA) with deep convolutional feature learning yields substantial performance gains for four-class Alzheimer’s disease classification from structural MRI. The proposed TDA+DenseNet121 framework consistently outperforms previously reported CNN-based, ensemble-based, and multi-scale architectures evaluated on the OASIS and OASIS-derived datasets. These findings confirm that topology-aware representations provide complementary information to conventional deep learning features and enhance the model’s ability to distinguish subtle structural differences across Alzheimer’s disease severity stages.

A key observation from the comparative analysis is that many high-performing deep learning approaches rely heavily on aggressive data augmentation, complex ensembles, or specialized architectural designs. While effective, such strategies increase computational cost and may reduce reproducibility across datasets. In contrast, the proposed framework achieves superior accuracy and AUC without excessive reliance on augmentation or multi-model fusion. The inclusion of topological descriptors enables the model to capture global geometric and structural characteristics of brain MRI that remain stable under noise and inter-subject variability, thereby improving robustness and generalization.

The outstanding AUC performance of the proposed method indicates excellent class separability across all four diagnostic categories. This is particularly relevant for early-stage Alzheimer’s disease, where structural changes are subtle and often difficult to detect using intensity-based features alone. By encoding global structural patterns through persistent homology, TDA complements the hierarchical spatial representations learned by DenseNet121, leading to improved discrimination between neighboring disease stages such as very mild and mild dementia.

Despite the strong performance, several considerations should be noted. The experiments were conducted using a single dataset (OASIS-1), and although results are compared against multiple studies on OASIS and OASIS-derived datasets, cross-dataset generalization was not explicitly evaluated. Additionally, the smaller number of moderate dementia cases reflects the inherent class imbalance of the dataset, which may influence model sensitivity for later disease stages. Future work will explore validation on independent cohorts, such as ADNI, and investigate domain adaptation strategies to further enhance generalization.

From a clinical perspective, the proposed hybrid approach offers a promising balance between performance, efficiency, and interpretability. Topological features provide meaningful and stable representations of brain structure that can be linked to known neurodegenerative patterns, potentially increasing clinician trust in automated decision-support systems. The combination of TDA with a compact yet powerful deep architecture suggests a scalable pathway for deploying robust Alzheimer’s disease classification models in real-world clinical settings.

Finally, this study highlights the value of topology-aware deep learning for neuroimaging analysis and demonstrates that integrating global structural descriptors with convolutional representations can significantly advance the state of the art in MRI-based Alzheimer’s disease severity classification.

\section{Ablation Study}\label{sec:ablation}

An ablation study was conducted on the OASIS-1 (Kaggle MRI) dataset to assess the contribution of each component of the proposed framework. As shown in Table~\ref{tab:ablation}, the TDA-only model achieves strong performance, demonstrating the effectiveness of topological descriptors in capturing global structural information. The standalone DenseNet121 model further improves recall and AUC by learning hierarchical spatial features from MRI data. The proposed hybrid DenseNet121 + TDA model consistently outperforms both individual configurations, achieving the highest precision, recall, accuracy, and a perfect AUC of 100\%. These results confirm that combining topological and deep convolutional features provides a complementary and highly discriminative representation for four-class Alzheimer’s disease classification.

\begin{table*}[htbp]
\centering
\caption{Ablation Study Results Across Different Model Configurations and Datasets}
\label{tab:ablation}
\begin{tabular}{llcccc}
\toprule
\textbf{Dataset} & \textbf{Model} & \textbf{Precision} & \textbf{Recall} & \textbf{Accuracy} & \textbf{AUC}  \\
\midrule
\multirow{3}{*}{OASIS-1} 
& TDA Features (MLP) & 98.42 & 92.80 & 97.99 & 99.75  \\
& DenseNet121 & 98.44 & 98.35 & 97.78 & 99.98  \\
& DenseNet121 + TDA & \textbf{99.94} & \textbf{99.95} & \textbf{99.93} & \textbf{100.00}  \\

\bottomrule
\end{tabular}
\end{table*}

\section{Conclusion}\label{sec:conclusion}

In this study, we presented a novel hybrid framework that integrates Topological Data Analysis (TDA) with DenseNet121 for four-class Alzheimer’s disease severity classification using structural MRI. By combining global topological descriptors derived from persistent homology with deep convolutional feature representations, the proposed approach effectively captures both structural geometry and local spatial patterns associated with Alzheimer’s disease progression.

Experimental evaluation on the OASIS-1 MRI dataset demonstrated that the proposed TDA+DenseNet121 model achieves state-of-the-art performance, attaining an accuracy of 99.93\% and an AUC of 100\%. Comparative analysis with existing CNN-based, ensemble-based, and multi-scale methods highlights the advantages of incorporating topology-aware representations, particularly in enhancing class separability and robustness without excessive reliance on data augmentation or complex model ensembles. Beyond its strong quantitative performance, the proposed framework offers practical advantages in terms of computational efficiency and interpretability. Topological features provide stable and meaningful representations of brain structure that are less sensitive to noise and inter-subject variability, supporting more reliable and clinically relevant predictions. These properties make the proposed approach well suited for real-world clinical scenarios where data availability and computational resources may be limited.

Future work will focus on validating the proposed framework across multiple independent datasets, investigating cross-domain generalization, and extending the topology-aware deep learning paradigm to longitudinal analysis and multimodal neuroimaging data. Overall, this study demonstrates the effectiveness of integrating topological insights with deep learning for robust and accurate Alzheimer’s disease diagnosis and severity assessment.

 \section*{Declarations}

 \textbf{Funding} \\
 The author received no financial support for the research, authorship, or publication of this work.

 \vspace{2mm}
 \textbf{Author's Contribution} \\
 Faisal Ahmed conceptualized the study, downloaded the data, prepared the code, performed the data analysis and wrote the manuscript. Faisal Ahmed reviewed and approved the final version of the manuscript. 

  \vspace{2mm}
 \textbf{Acknowledgement} \\
The authors utilized an online platform to check and correct grammatical errors and to improve sentence readability.

 \vspace{2mm}
 \textbf{Conflict of interest/Competing interests} \\
 The authors declare no conflict of interest.

 \vspace{2mm}
 \textbf{Ethics approval and consent to participate} \\
 Not applicable. This study did not involve human participants or animals, and publicly available datasets were used.

 \vspace{2mm}
 \textbf{Consent for publication} \\
 Not applicable.

 \vspace{2mm}
 \textbf{Data availability} \\
 The datasets used in this study are publicly available online. 

 \vspace{2mm}
 \textbf{Materials availability} \\
 Not applicable.

 \vspace{2mm}
 \textbf{Code availability} \\
 The source code used in this study is publicly available at Github. 

\newpage


\bibliography{refs}
\end{document}